\documentclass[8pt]{article}
\usepackage{spconf,amsmath,graphicx}
\usepackage{epsfig}
\usepackage{subfigure}
\usepackage{algorithm}
\usepackage{algorithmic}
\usepackage{amsthm}
\usepackage{multicol}
\usepackage{multirow}
\usepackage{array}
\usepackage{pslatex}
\usepackage{times,enumerate}
\usepackage[small]{caption}
\usepackage{placeins}
\usepackage[utf8]{inputenc} 
\usepackage[T1]{fontenc}    
\usepackage{url}       
\usepackage{amsfonts}       
\usepackage{amsmath}        
\usepackage{amssymb}
\usepackage{amstext}
\usepackage{amsthm}
\usepackage{graphicx}

\DeclareMathOperator*{\argmin}{arg\,min}

\newcommand{\measurement}{\mathbf{y}}
\newcommand{\signal}{\mathbf{x}}
\newcommand{\reps}{\mathbf{c}} 
\newcommand{\repsHat}{\hat{\reps}} 

\title{A Deep Learning Approach to Block-based Compressed Sensing of Images}
\name{Amir Adler$^{\star}$\thanks{This research is partly supported by ERC Grant agreement no. 320649, and partly by the Intel Collaborative Research Institute for Computational Intelligence (ICRI-CI).}, David Boublil$^{\dagger}$, Michael Elad $^{\star}$, Michael Zibulevsky$^{\star}$}

\address{$^{\star}$ Computer Science Department, Technion, Haifa 32000, Israel \\
    $^{\dagger}$ Electrical Engineering Department, Technion, Haifa 32000, Israel}
\begin{document}
%
\maketitle
\begin{abstract}
Compressed sensing (CS) is a signal processing framework for efficiently reconstructing a signal from a small number of measurements, obtained by linear projections of the signal. Block-based CS is a lightweight CS approach that is mostly suitable for processing very high-dimensional images and videos: it operates on local patches, employs a low-complexity reconstruction operator and requires significantly less memory to store the sensing matrix. In this paper we present a deep learning approach for block-based CS, in which a fully-connected network performs both the block-based linear sensing and non-linear reconstruction stages. During the training phase, the sensing matrix and the non-linear reconstruction operator are \emph{jointly} optimized, and the proposed approach outperforms state-of-the-art both in terms of reconstruction quality and computation time. For example, at a $25\%$ sensing rate the average PSNR advantage is 0.77dB and computation time is over 200-times faster.
\end{abstract}

\section{Introduction}
\label{sec:intro}
\vskip -0.25cm
Compressed sensing \cite{Donoho2006,Candes2008} is a mathematical framework that defines the conditions and tools for the recovery of a signal from a small number of its linear projections (i.e. measurements). In the CS framework, the measurement device acquires the signal in the linear projections domain, and the full signal is reconstructed by convex optimization techniques. CS has diverse applications including image acquisition \cite{Romberg2008}, radar imaging \cite{5420035}, Magnetic Resonance Imaging (MRI) \cite{4472246, 6153065}, spectrum sensing \cite{6179814}, indoor positioning \cite{6042868}, bio-signals acquisition \cite{6184345}, and sensor networks \cite{6159081}. In this paper we address the problem of block-based CS (BCS) \cite{Fowler-NOW}, which employs CS on distinct low-dimensional segments of a high-dimensional signal. BCS is mostly suitable for processing very high-dimensional images and video, where it operates on distinct local patches. Our approach is based on a deep neural network \cite{Bengio-2009}, which simultaneously learns the linear sensing matrix and the non-linear reconstruction operator.\\
The contributions of this paper are two-fold: (1) It presents for the first time, to the best knowledge of the authors, the utilization of a fully-connected deep neural network for the task of BCS; and (2) The proposed network performs both the linear sensing and non-linear reconstruction operators, and during training these operators are \emph{jointly} optimized, leading to a significant advantage compared to state-of-the-art.\\
This paper is organized as follows: section \ref{Problem Formulation} introduces CS concepts, and motivates the utilization of BCS for very high-dimensional images and video. Section \ref{The Proposed Approach} presents the deep neural network approach, and discusses structure and training aspects. Section \ref{Results} evaluates the performance of the proposed approach for compressively sensing and reconstructing natural images, and compares it with state-of-the-art BCS methods and full-image Total Variation-based CS. Section \ref{Conclusions} concludes the paper and discusses future research directions. \vskip -0.25cm
\section{Compressed Sensing Overview}
\vskip -0.25cm
\label{Problem Formulation}
\subsection{Full-Signal Compressed Sensing}
Given a signal $\signal \in \mathbf{R}^N$, an $M \times N$ sensing matrix $\Phi$ (such that $M \ll N$) and a measurements vector $\measurement = \Phi \signal$, the goal of CS is to recover the signal from its measurements. The sensing rate is defined by $R=M/N$, and since $R \ll 1$ the recovery of $\signal$ is not possible in the general case. According to CS theory \cite{Donoho2006,Candes2008}, signals that have a sparse representation in the domain of some linear transform can be exactly recovered with high probability from their measurements: let $\signal = \Psi \reps $, where $\Psi$ is the inverse transform, and $\reps$ is a sparse coefficients vector with only $S \ll N$ non-zeros entries, then the recovered signal is synthesized by $ \hat{\signal} = \Psi \repsHat$, and $\repsHat$ is obtained by solving the following convex optimization program:
\begin{equation}
 \repsHat  = \argmin_{\reps{'}} \left\|\reps{'}\right\|_1 \text{ subject to } \measurement = \Phi \Psi \reps{'},
\end{equation}
\vskip -0.25cm
\noindent where $\left\|\alpha\right\|_1$ is the $l_1$-norm, which is a convex relaxation of the $l_0$ pseudo-norm that counts the number of non-zero entries of $\alpha$. The exact recovery of $\signal$ is guaranteed with high probability if $\reps$ is sufficiently sparse and if certain conditions are met by the sensing matrix and the transform.

\begin{table*}[]
  \caption{Average reconstruction PSNR [dB] and SSIM vs. sensing rate (R=M/N): for each method and sensing rate, the result is displayed as PSNR | SSIM  (each result is the average over the 10 test images).}
  \label{Reconstruction-Quality}
  \centering
\resizebox{\textwidth}{!}{
  \begin{tabular}{lccccc}
   \hline
    Method                        & R = 0.1  & R = 0.15 & R = 0.2 & R = 0.25 &  R = 0.3\\
   \hline
   Proposed (block-size = 16$\times$16)     & \textbf{28.21} | \textbf{0.916} & \textbf{29.73} | \textbf{0.948} & \textbf{31.03} | \textbf{0.965}  & \textbf{32.15} | \textbf{0.976} & \textbf{33.11} | \textbf{0.983}\\
   BCS-SPL-DDWT (16$\times$16)\cite{Fowler2009}  & 24.92 | 0.789 & 26.12 | 0.834 & 27.17 | 0.873 & 28.16 | 0.898 & 29.02 | 0.917 \\
   BCS-SPL-DDWT (32$\times$32)\cite{Fowler2009}   & 24.99 | 0.781 & 26.40 | 0.833 & 27.46 | 0.868 & 28.43 | 0.894 & 29.29 | 0.914 \\
   MH-BCS-SPL (16$\times$16)  \cite{MH-Fowler}    & 26.01 | 0.827 & 27.92 | 0.888 & 29.46 | 0.919 & 30.69 | 0.939 & 31.69 | 0.952 \\
   MH-BCS-SPL (32$\times$32)  \cite{MH-Fowler}    & 26.79 | 0.845 & 28.51 | 0.895 & 29.81 | 0.923 & 30.77 | 0.938 & 31.73 | 0.950 \\
   MS-BCS-SPL \cite{MS-Fowler}                      & 27.32 | 0.883 & 28.77 | 0.909 & 30.04 | 0.934 & 31.15 | 0.956 & 32.05 | 0.974 \\
   MH-MS-BCS-SPL \cite{MH-Fowler}                     & 27.74 | 0.889 & 29.10 | 0.919 & 30.78 | 0.947 & 31.38 | 0.960 & 32.82 | 0.979 \\
   TV (Full Image) \cite{Romberg2008}               & 27.41 | 0.867 & 28.57 | 0.890 & 29.62 | 0.909 & 30.63 | 0.926 & 31.59 | 0.939 \\
   \hline
  \end{tabular}}
\end{table*}
\subsection{Block-based Compressed Sensing}
\label{Block-based Compressed Sensing}
Consider applying CS to an image of $L \times L$ pixels: the technique described above can be employed by column-stacking (or row-stacking) the image to a vector $\signal \in \mathbb{R}^{L^2}$, and the dimensions of the measurement matrix $\Phi$ and the inverse transform $\Psi$ are $M \times {L^2}$ and $ {L^2} \times {L^2}$, respectively. For modern high-resolution cameras, a typical value of $L$ is in the range of $2000$ to $4000$, leading to overwhelming memory requirements for storing $\Phi$ and $\Psi$: for example, with $L=2000$ and a sensing rate $R=0.1$ the dimensions of $\Phi$ are $400,000 \times 4,000,000$ and of $\Psi$ are $4,000,000 \times 4,000,000$. In addition, the computational load required to solve the CS reconstruction problem becomes prohibitively high. Following this line of arguments, a BCS framework was proposed in \cite{Gan2007}, in which the image is decomposed into non-overlapping blocks (i.e. patches) of $B \times B$ pixels, and each block is compressively sensed independently. The full-size image is obtained by placing each reconstructed block in its location within the reconstructed image canvas, followed by full-image smoothing. The dimensions of the block sensing matrix $\Phi_B$ are ${B^2}R\times{B^2}$, and the measurement vector of the \emph{i}-th block is given by:
\begin{equation}
\label{block-based sensing}
\measurement_i = \Phi_B \signal_i,
\end{equation}
\noindent where $\signal_i \in \mathbb{R}^{B^2}$ is the column-stacked block, and $\Phi_B$ was chosen in \cite{Gan2007} as an orthonormalized i.i.d Gaussian matrix. Following a per-block minimum mean squared error reconstruction stage, a full-image iterative hard-thresholding algorithm is employed for improving full-image quality. An improvement to the performance of this approach was proposed by \cite{Fowler2009}, which employed the same BCS approach as \cite{Gan2007} and evaluated the incorporation of directional transforms such as the Contourlet Transform (CT) and the Dual-tree Discrete Wavelet Transform (DDWT) in conjunction with a Smooth Projected Landweber (SPL) \cite{bertero1998introduction} reconstruction of the full image. The conclusion of the experiments conducted in \cite{Fowler2009} was that in most cases the DDWT transform offered the best performance, and we term this method as BCS-SPL-DDWT. A multi-scale approach was proposed by \cite{MS-Fowler}, termed MS-BCS-SPL, which improved the performance of BCS-SPL-DDWT by applying the block-based sensing and reconstruction stages in multiple scales and sub-bands of a discrete wavelet transform. A different block dimension was employed for each scale and with a 3-level transform, dimensions of $B=64,32,16$ were set for the high, medium and low resolution scales, respectively. A multi-hypothesis approach was proposed in \cite{MH-Fowler} for images and videos, which is suitable for either spatial domain BCS (termed MH-BCS-SPL) or multi-scale BCS (termed MH-MS-BCS-SPL). In this approach, multiple predictions of a block are computed from neighboring blocks in an initial reconstruction of the full image, and the final prediction of the block is obtained by an optimal linear combination of the multiple predictions. For video frames, previously reconstructed adjacent frames provide the sources for multiple predictions of a block. The multi-scale version of this approach provides the best performance among all above mentioned BCS methods. A detailed survey of BCS theory and performance is provided in \cite{Fowler-NOW}, which describes additional applications such as BCS of multi-view images and video, and motion-compensated BCS of video.
\section{The Proposed Approach}
\vskip -0.25cm
\label{The Proposed Approach}
In this paper we propose to employ a deep neural network that performs BCS by processing each block independently\footnote{In this paper we treat only block-based processing, and a full-image post-processing stage is not performed.} as described in section \ref{Block-based Compressed Sensing}. Our choice is motivated by the outstanding success of deep neural networks for the task of full-image denoising \cite{Burger} in which a 4-layer neural network achieved state-of-the-art performance by block-based processing. In our approach, the first hidden layer performs the linear block-based sensing stage (\ref{block-based sensing}) and the following hidden layers perform the non-linear reconstruction stage. The advantage and novelty of this approach is that during training, the sensing matrix and the non-linear reconstruction operator are \emph{jointly} optimized, leading to better performance than state-of-the-art at a fraction of the computation time. \\ The proposed fully-connected network includes the following layers: (1) an input layer with $B^2$ nodes; (2) a compressed sensing layer with $B^{2}R$ nodes, $R\ll1$ (its weights form the sensing matrix); (3) $K\ge1$ reconstruction layers with $B^{2}T$ nodes, each followed by a ReLU \cite{icml2010_NairH10} activation unit, where $T>1$ is the redundancy factor; and (4) an output layer with $B^2$ nodes. Note that the performance of the network depends on the block-size $B$, the number of reconstruction layers $K$, and their redundancy $T$. We have evaluated\footnote{The network was implemented using Torch7 \cite{Collobert_NIPSWORKSHOP_2011} scripting language, and trained on NVIDIA Titan X GPU card.} these parameters by a set of experiments that compared the average reconstruction PSNR of 10 test images, depicted in Figure \ref{test_images}, and by training the network with 5,000,000 distinct patches, randomly extracted from 50,000 images in the LabelMe dataset \cite{LabelMe}. The chosen optimization algorithm was AdaGrad \cite{AdaGrad} with a learning rate of $0.005$ (100 epochs), and batch size of 16. Our study revealed that best\footnote{Note that by increasing significantly the training set, slightly different values of $B$, $K$, and $T$ may provide better results, as discussed in \cite{Burger}.} performance were achieved with a block size $B\times B = 16\times16$, $K=2$ reconstruction layers and redundancy $T=8$, leading to a total of 4,780,569 parameters ($R=0.1$). Table \ref{Reconstruction-Quality-vs-Size} provides a comparison for varying the block size between $8\times8$ to $20\times20$ (with 2 reconstruction layers and redundancy of 8), and indicates that block size of $16\times16$ provides the best results. Table \ref{Reconstruction-Quality-vs-Redundancy} provides a comparison for varying the redundancy between 2 to 12 (with 2 reconstruction layers and block size of $16\times 16$), and indicates that a redundancy of 8 provides the best results. Table \ref{Reconstruction-Quality-vs-Layers} provides a comparison for varying the number of hidden reconstruction layers between 1 to 8 (with redundancy of 8 and block size of $16\times 16$), and indicates that two reconstruction layers provided the best performance.
\begin{table}[]
 \vskip -0.25cm
  \caption{Reconstruction PSNR [dB] vs. block size ($B\times B$)}
  \label{Reconstruction-Quality-vs-Size}
  \centering
  \begin{tabular}{cllll}
    \hline
    Training Examples & $B=8$        & $B=12$     & $B=16$      & $B=20$ \\
    \hline
    $5 \times 10^6$   &     27.21  &  27.66   & \textbf{28.21}   &  27.73 \\
    \hline
  \end{tabular}
\end{table}
\begin{table}[]
 \vskip -0.25cm
  \caption{Reconstruction PSNR [dB] vs. network redundancy}
  \label{Reconstruction-Quality-vs-Redundancy}
  \centering
  \begin{tabular}{cllll}
    \hline
    Training Examples & $T=2$ & $T=4$ & $T=8$ & $T=12$ \\
    \hline
     $5 \times 10^6$   & 27.99  &  28.11 & \textbf{28.21}   &  28.15 \\
    \hline
  \end{tabular}
\end{table}
\begin{table}[htb]
 \vskip -0.25cm
  \caption{Reconstruction PSNR [dB] vs. no. of reconstruction layers}
  \label{Reconstruction-Quality-vs-Layers}
  \centering
  \begin{tabular}{ccccc}
    \hline
    Training Examples & $K=1$ & $K=2$ & $K=4$ & $K=8$ \\
    \hline
     $5 \times 10^6$   & 27.98  &  \textbf{28.21} & 28.18   &  27.07 \\
    \hline
  \end{tabular}
\end{table}
\begin{table}[t]
  \caption{Computation time at R=0.25 ($512 \times 512$ images):}
  \label{Comp-Time}
   \centering
  \begin{tabular}{lc}
  \hline
    Method    &  Time [seconds]\\
     \hline
    Proposed  & 0.80\\
    BCS-SPL-DDWT ($16 \times 16$) \cite{Fowler2009}   & 13.57\\
    BCS-SPL-DDWT ($32 \times 32$) \cite{Fowler2009}   & 13.10\\
    MH-BCS-SPL ($16 \times 16$) \cite{MH-Fowler}   &  144.61\\
    MH-BCS-SPL ($32 \times 32$) \cite{MH-Fowler}   &  69.73\\
    MS-BCS-SPL \cite{MS-Fowler}   & 6.39\\
    MH-MS-BCS-SPL \cite{MH-Fowler} & 207.32\\
    TV (Full Image) \cite{Romberg2008} & 1675.09\\
    \hline
  \end{tabular}
\end{table}
\section{Performance Evaluation}
\begin{figure*}[]
\begin{minipage}{\linewidth}
\vskip -1cm
 \makebox[\linewidth]{
\centering
\includegraphics[width=200mm, scale=0.55]{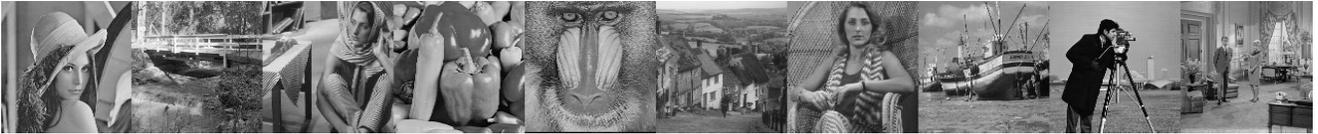}}
\vskip -1cm
 \caption{Test images ($512 \times 512$): 'lena', 'bridge', 'barbara', 'peppers', 'mandril', 'houses', 'woman', 'boats', 'cameraman' and 'couple'.}
\label{test_images}
 \end{minipage}
\end{figure*}

\begin{figure*}[]
 \begin{minipage}{\linewidth}
 \vskip -1cm
 \makebox[\linewidth]{
\centering
\includegraphics[width=200mm,scale=0.5]{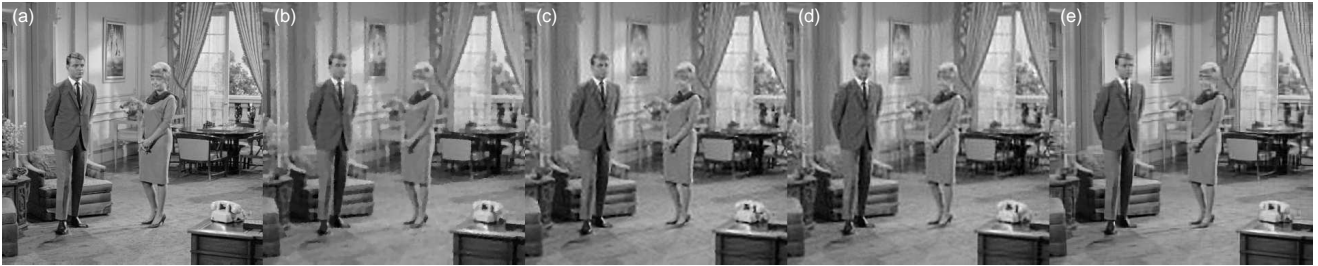}}
\vskip -1cm
  \caption{Reconstruction of 'couple' at R = 0.1 (PSNR [dB] | SSIM): (a) Original; (b) Full image TV (27.1691 | 0.8812); (c) MS-BCS-SPL (26.8429 | 0.8756); (d) MH-MS-BCS-SPL (27.1804 | 0.8877); and (e) Proposed (28.5902 | 0.9414).}
 \label{results_comp_01}
 \end{minipage}
\end{figure*}

\begin{figure*}[]
 \begin{minipage}{\linewidth}
 \vskip -1cm
 \makebox[\linewidth]{
\centering
\includegraphics[width=200mm,scale=0.5]{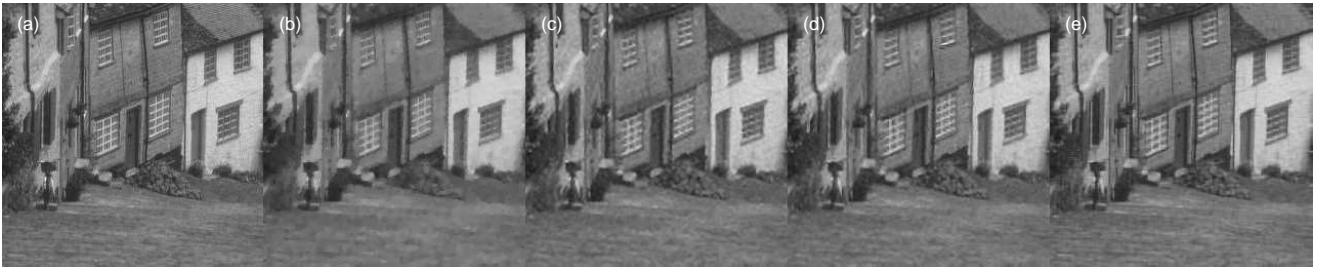}}
\vskip -1cm
 \caption{Reconstruction of 'houses' at R = 0.2 (PSNR [dB] | SSIM): (a) Original; (b) Full image TV (31.0490 | 0.9304); (c) MS-BCS-SPL (31.4317 | 0.9497); (d) MH-MS-BCS-SPL (31.7030 | 0.9544); and (e) Proposed (32.9328 | 0.9766).}
 \label{results_comp_02}
  \end{minipage}
\end{figure*}

\begin{figure*}[]
 \begin{minipage}{\linewidth}
 \vskip -1cm
 \makebox[\linewidth]{
\centering
\includegraphics[width=200mm,scale=0.5]{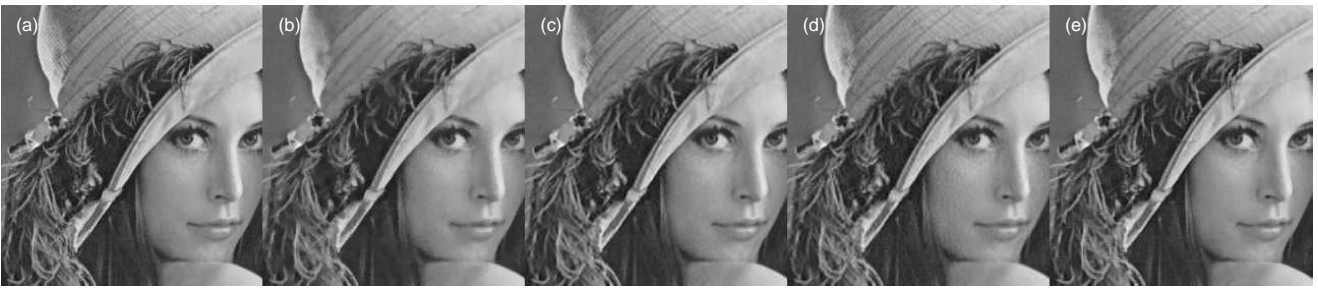}}
\vskip -1cm
 \caption{Reconstruction of 'lena' at R = 0.25 (PSNR [dB] | SSIM): (a) Original; (b) Full image TV (35.4202 | 0.9718); (c) MS-BCS-SPL (36.5555 | 0.9861); (d) MH-MS-BCS-SPL (35.7346 | 0.9825; and (e) Proposed (36.3734 | 0.9910).}
 \label{results_comp_025_1}
  \end{minipage}
\end{figure*}

\begin{figure*}[]
 \begin{minipage}{\linewidth}
 \vskip -1cm
 \makebox[\linewidth]{
\centering
\includegraphics[width=200mm,scale=0.5]{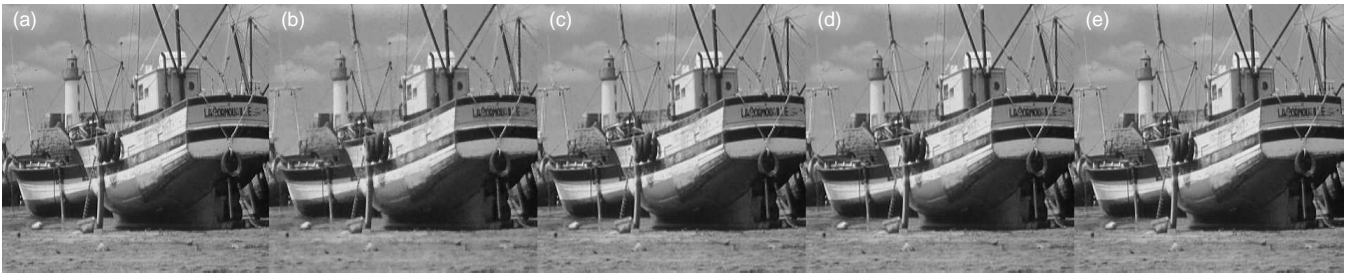}}
\vskip -1cm
 \caption{Reconstruction of 'boats' at R = 0.3 (PSNR[dB] | SSIM): (a) Original; (b) Full image TV (32.5986 | 0.9598); (c) MS-BCS-SPL (32.5063 | 0.9835); (d) MH-MS-BCS-SPL (32.7697 | 0.9847); and (e) Proposed (34.0065 | 0.9914).}
 \label{results_comp_025_2}
  \end{minipage}
\end{figure*}

\vskip -0.25cm
\label{Results}
This section provides performance evaluation results of the proposed approach\footnote{A MATLAB package implementing the proposed approach is available at: \url{http://www.cs.technion.ac.il/~adleram/BCS_DNN_2016.zip}} vs. the leading BCS approaches: BCS-SPL-DDWT \cite{Fowler2009}, MS-BCS-SPL \cite{MS-Fowler}, MH-BCS-SPL \cite{MH-Fowler} and MH-MS-BCS-SPL \cite{MH-Fowler}, using the original code provided by their authors. The proposed approach was employed with block size $16 \times 16$, BCS-SPL-DDWT with block sizes of $16 \times 16$ and $32 \times 32$ (the optimal size for this method), MH-BCS-SPL with block sizes of $16 \times 16$ and $32 \times 32$ (the optimal size for this method). MS-BCS-SPL and MH-MS-BCS-SPL utilized a 3-level discrete wavelet transform with block sizes as indicated in section \ref{Block-based Compressed Sensing} (their optimal settings). In addition, we also compared to the classical full-image Total Variation (TV) CS approach of \cite{Romberg2008} that utilizes a sensing matrix with elements from a discrete cosine transform and Noiselet vectors. Reconstruction performance was evaluated for sensing rates in the range of $R=0.1$ to $R=0.3$, using the average PSNR and SSIM \cite{SSIM} over the collection of 10 test images ($512 \times 512$ pixels), depicted in Figure \ref{test_images}. Reconstruction results are summarized in Table \ref{Reconstruction-Quality}, and reveal a consistent advantage of the proposed approach vs. all BCS methods as well as the full-image TV approach. Visual quality comparisons (best viewed in the electronic version of this paper) are provided in Figures \ref{results_comp_01}-\ref{results_comp_025_2}, and demonstrate the high visual quality of the proposed approach. Computation time comparison at a sensing rate $R=0.25$, with a MATLAB implementation of all tested methods (without GPU), is provided in Table \ref{Comp-Time} and demonstrates that the proposed approach is over 200-times faster than state-of-the-art (MH-MS-BCS-SPL), and over 1600-times faster than full-image TV CS.
\section{Conclusions}
\vskip -0.25cm
\label{Conclusions}
This paper presents a deep neural network approach to BCS, in which the sensing matrix and the non-linear reconstruction operator are jointly optimized during the training phase. The proposed approach outperforms state-of-the-art both in terms of reconstruction quality and computation time, which is two orders of magnitude faster than the best available BCS method. Our approach can be further improved by extending it to compressively sense blocks in a multi-scale representation of the sensed image, either by utilizing standard transforms or by deep learning of new transforms using convolutional neural networks.
\vskip -0.25cm
\bibliographystyle{IEEEbib}
\bibliography{bcs_dnn_2016_Bibliography}

\end{document}